\documentclass[letterpaper, 10 pt, conference]{ieeeconf}

\IEEEoverridecommandlockouts
\usepackage{graphicx}
\usepackage{amsmath}
\usepackage{amssymb}
\usepackage{amsfonts}
\usepackage{booktabs}
\usepackage{multirow}
\usepackage{subcaption}
\usepackage{cite}
\usepackage{url}

\graphicspath{{Figures/}}

\newcommand{\Expectation}[1]{\mathop{\mathbb E}\limits_{#1}}

\newtheorem{theorem}{Theorem}

\title{\LARGE \bf
SCQ: Stabilizing Conservative Q-Learning with Sigmoid-Bounded Entropy
}

\author{\authorblockN{Xiefeng Wu\authorrefmark{1}, Shu Zhang\authorrefmark{2}, Zhaojie Chu\authorrefmark{3}, and Mingyu Hu\authorrefmark{2}\thanks{Corresponding author: Mingyu Hu (email: mingyuhu@whu.edu.cn).}}
\authorblockA{\authorrefmark{1}School of Computer Science, Wuhan University, Wuhan, China\\
Email: wuxiefeng@whu.edu.cn}
\authorblockA{\authorrefmark{2}School of Electronic Information, Wuhan University, Wuhan, China\\
Email: 00033521@whu.edu.cn, mingyuhu@whu.edu.cn}
\authorblockA{\authorrefmark{3}School of Internet, Anhui University, Hefei, China\\
Email: zjchu\_china@163.com}}

\begin{document}

\maketitle
\thispagestyle{empty}
\pagestyle{empty}

\begin{abstract}
Offline-to-online reinforcement learning reduces interaction cost for real-world robot learning but suffers from persistent value estimation instability. Existing methods address this through pessimistic regularization, lower-bound calibration, and architectural normalization, but an overlooked source of instability lies in the entropy formulation: the standard log-entropy term can become negative, destabilizing policy updates. We introduce SCQ (Sigmoid-Bounded Conservative Q-Learning), which replaces this term with a sigmoid-bounded formulation that stays strictly positive. SCQ retains conservative Q regularization and return-based lower-bound calibration, stabilizing policy optimization without sacrificing exploration. We evaluate SCQ on D4RL (Minari) benchmarks under both single-demonstration and standard dataset settings, as well as on simulation and real-world visual tasks. SCQ matches or exceeds baseline performance while exhibiting more stable training dynamics across state-based and visual benchmarks, and transfers to four real-robot platforms including manipulation, wheeled, quadruped, and humanoid systems. A direct clipping intervention that removes negative log-probability contributions, together with gradient-matched positive-score controls, indicates that positivity rather than a particular score shape alone drives much of the improvement. Project website: \url{https://scq-rl.github.io}.
\end{abstract}

\begin{figure*}[!t]
  \centering
  \includegraphics[width=\textwidth]{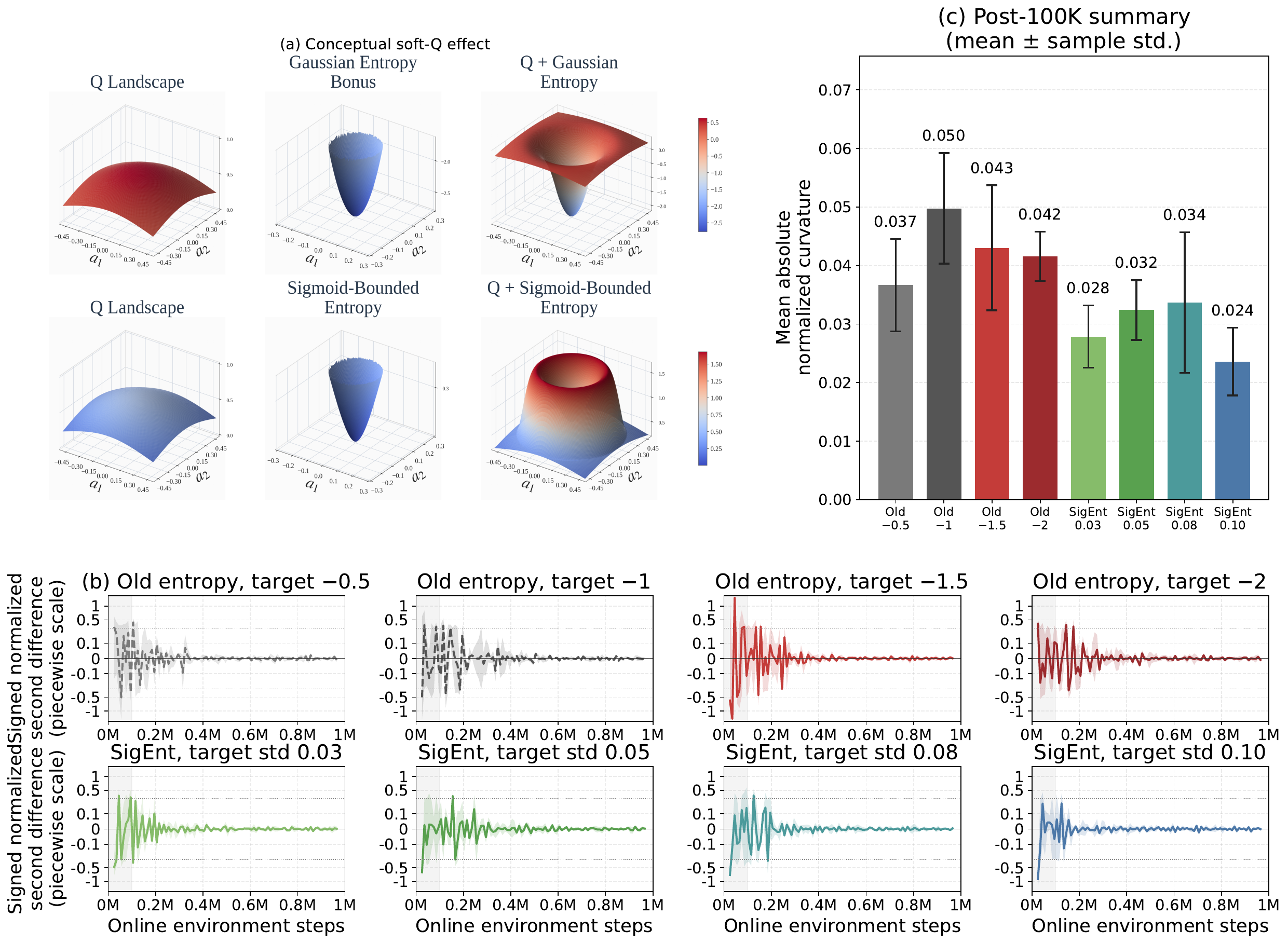}
  \caption{Negative-entropy mechanism and Q-gradient temporal-variability diagnostic. \textbf{(a)} Conceptual local soft-$Q$ effect for a Gaussian policy with $\sigma_\pi=0.1$ and actions sampled within $1.5\sigma_\pi$ of the policy mean. The standard sample-wise entropy contribution can lower the entropy-augmented objective near the current policy mean, whereas the sigmoid-bounded score remains strictly positive and bounded. \textbf{(b)} Door Human one-shot runs without pretraining. Each curve is the five-seed mean of the signed normalized temporal second difference of the logged batch mean $g_Q=\|\nabla_{\mu}(-Q)\|_2$, computed from 10K-step bins; shaded bands show $\pm$ sample standard deviation. The rows cover old entropy targets $-0.5$, $-1$, $-1.5$, and $-2$, and SigEnt target standard deviations $0.03$, $0.05$, $0.08$, and $0.10$. Curves use common complete 10K-step bins through 1M. The vertical axis uses a symmetric piecewise scale that expands the central $[-0.2,0.2]$ range; the shaded vertical interval marks the initial 0--100K transient. \textbf{(c)} Mean absolute normalized temporal second difference after 100K steps. Bars show the five-seed mean and whiskers show $\pm$ sample standard deviation ($n=5$), computed separately within each seed before aggregation.}
  \label{fig:entropy_mechanism_gradient}
\end{figure*}

\section{Introduction}

Reinforcement learning (RL) is a promising approach to robot skill acquisition, but learning directly in the real world is constrained by expensive interaction and unstable policy optimization. Offline-to-online RL reduces this cost by initializing learning from demonstrations and then improving the policy through online interaction. Existing methods such as Cal-QL~\cite{calql} and RLPD~\cite{rlpd} stabilize value learning through conservative Q regularization, return-based lower-bound calibration, and LayerNorm.

Despite these stabilization techniques, policy optimization can still exhibit oscillations and learning failures in one-shot and other limited-replay offline-to-online regimes. We identify a contributing factor that has not been addressed by prior work: the entropy formulation itself. In maximum-entropy actor-critic methods, the standard sample-wise entropy contribution can become negative for tanh-squashed Gaussian policies. When added to the Q-value, these negative contributions locally depress the entropy-augmented objective around sampled policy actions and can change the magnitude and direction of the policy-mean update. The resulting policy movement changes the actions at which the critic is queried; changes in those value estimates then alter the next policy update. Under limited replay, this feedback can amplify policy-mean oscillation and destabilize value estimation.

We introduce \textbf{SCQ} (Sigmoid-Bounded Conservative Q-Learning), a stability-oriented offline-to-online RL framework built on Cal-QL. The central change in SCQ is a strictly positive, sigmoid-bounded entropy formulation that replaces the standard sample-wise entropy contribution. Its positive bounded range prevents sign reversal of the entropy contribution and limits its distortion of the effective Q landscape, reducing policy-mean oscillation and the resulting variation in critic queries. In the offline-to-online setting, SCQ retains Cal-QL's conservative objective and return-based lower-bound calibration; critic-only LayerNorm further stabilizes value optimization. Our empirical hypothesis is that preventing negative entropy contributions improves learning relative to the standard entropy formulation. We test it directly by clipping only the negative branch of standard log-probability entropy and separately compare against locally gradient-matched ReLU and Softplus scores.

We evaluate SCQ on benchmark and real-world tasks. Tasks range from standard full-dataset benchmarks to one-shot learning from a single demonstration, and include four real-robot platforms spanning manipulation, wheeled, quadruped, and humanoid systems.

Our main contributions are as follows:
\begin{itemize}
    \item \textbf{Sigmoid-Bounded Entropy for Conservative Q-Learning.} We introduce a strictly positive, sigmoid-bounded entropy formulation for SAC-style actor-critic methods. It removes negative entropy contributions from both actor and critic updates, reducing policy-mean oscillation and subsequent value-estimation instability in limited-replay regimes.

    \item \textbf{Controlled mechanism tests and broad evaluation.} A direct clipped-log-probability intervention isolates the negative branch, while matched positive-score controls and training-dynamics diagnostics test the associated stability mechanism. We further evaluate SCQ in standard and one-shot benchmarks and across four real-world visual-control settings.
\end{itemize}

\section{Related Work}

We situate SCQ within three lines of research: offline-to-online RL methods that warmstart policy learning, entropy-regularized policy optimization that balances exploration and exploitation, and reinforcement learning deployed on physical robots.

\subsection{Offline-to-Online Reinforcement Learning}

Offline-to-online RL warmstarts online learning from offline data before further interaction. A common approach is to initialize the policy with offline RL methods such as IQL~\cite{iql}, CQL~\cite{cql}, AWAC~\cite{nair2020awac}, and TD3+BC~\cite{td3bc}, while Cal-QL~\cite{calql} is designed to facilitate this transition through conservative Q regularization and return-based calibration. RLPD~\cite{rlpd} alleviates Q-function overestimation through a larger Q-ensemble and critic-only LayerNorm. However, comparatively fewer studies evaluate these methods in few-shot, vision-based real-world scenarios.

\subsection{Entropy-Regularized Policy Optimization}

Maximum-entropy RL, exemplified by SAC~\cite{sac}, augments return with policy entropy. Prior work modifies entropy for sparse control, action allocation, expressive policies, or long-horizon objectives~\cite{tsallis2024,ace2024,s2ac2024,dacer2024,mind2025}. In contrast, SCQ addresses potentially negative sample-wise entropy contributions in conservative offline-to-online learning.

\subsection{Reinforcement Learning in Real-World Robot Settings}

Real-world RL must cope with limited interaction budgets, noisy observations, and safety constraints. One line of work addresses these challenges through system-level design: HIL-SERL incorporates human interventions into an off-policy loop~\cite{luo2024hilserl}, and RL-100 engineers a scalable real-world training pipeline~\cite{rl-100}. A second line leverages pretrained VLA policies as behavioral priors for online RL\@. RLT extracts a compact representation from a frozen VLA and trains a lightweight actor-critic on it~\cite{xu2026rltoken}; EXPO-FT directly fine-tunes the VLA together with an edit policy using the EXPO algorithm~\cite{dong2026expoft}. Both report clear improvements on several precise manipulation tasks, yet their generalization evidence is confined to a small set of single-arm settings with task-specific demonstrations and human supervision. Transfer across different embodiments and broader generalization remain unestablished. We attribute this primarily to value-estimation instability in the RL solvers they employ, which may further limit adaptation in low-data or out-of-distribution regimes. In contrast, this work targets a complementary setting: stabilizing the Q-learning solver itself so that reliable adaptation is possible with minimal demonstrations, no pretrained policy backbone, and diverse embodiments.

\begin{figure*}[!t]
\centering
  \begin{minipage}[c]{0.56\textwidth}
    \centering
    \includegraphics[trim=0 0 20 0,clip,width=\linewidth]{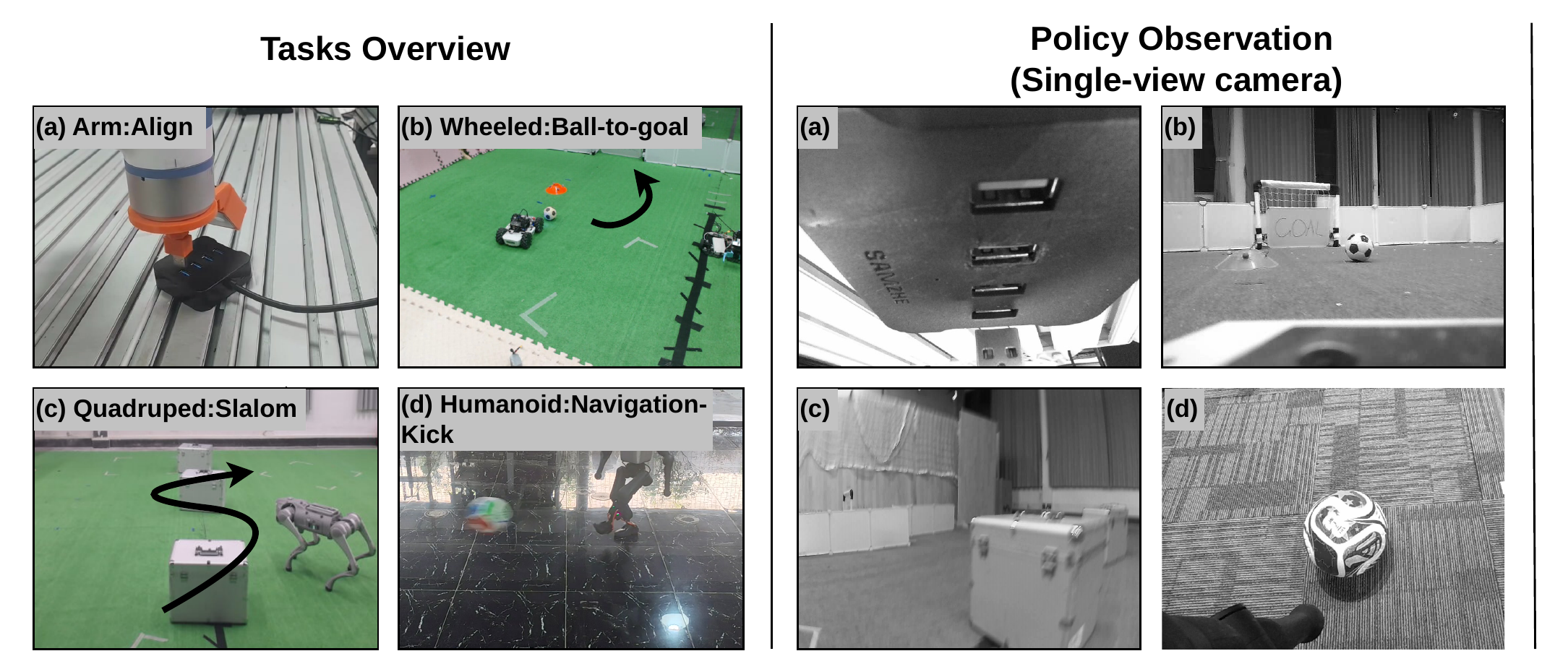}
  \end{minipage}\hfill
  \begin{minipage}[c]{0.42\textwidth}
    \centering
    {\footnotesize
    \setlength{\tabcolsep}{2pt}
    \renewcommand{\arraystretch}{1.12}
    \begin{tabular*}{\linewidth}{@{\extracolsep{\fill}}lrrrr@{}}
    \toprule
    Task & Online & Success & Policy & Expert \\
     & budget & rate & steps & steps \\
    \midrule
    \shortstack[l]{High-Precision\\Align} & 20K & 10/10 & -- & -- \\
    Ball-to-Goal & 30K & 6/10 & 3 & 26 \\
    Quadruped Slalom & 25K & 8/10 & 23 & 38 \\
    \shortstack[l]{Humanoid\\Navigation-Kick} & 150K & 10/10 & 210 & 313 \\
    \bottomrule
    \end{tabular*}}
  \end{minipage}
  \caption{Real-world task diversity and deployment summary. \textbf{Left:} High-Precision Align, Wheeled Ball-to-Goal, Quadruped Slalom, and Humanoid Navigation-Kick, using single-view local observations (grayscale for the first three tasks and RGB for humanoid navigation). \textbf{Right:} Online budget is the maximum permitted number of environment interactions. Success rate reports successful evaluation episodes out of 10 trials. For tasks with trajectory-level references, policy and expert columns report representative completion steps under the same inference framework. High-Precision Align is trained without an expert demonstration and evaluated by pose-error tolerances rather than trajectory completion, so both step entries are marked ``--''.}
  \label{fig:real_world_generalization}
\end{figure*}

\section{Notation}

We consider an MDP
$\mathcal{M}=\langle \mathcal{S},\mathcal{A},P,r,\gamma,\rho_0\rangle$,
where \(\mathcal{S}\) and \(\mathcal{A}\) are the state and action spaces, \(P(r,s',d_{\mathrm{term}}\mid s,a)\) is the transition distribution, \(r\) is the reward, \(d_{\mathrm{term}}\in\{0,1\}\) is the terminal indicator, \(\gamma\in(0,1)\) is the discount factor, and \(\rho_0\) is the initial-state distribution. We use \(\mathcal{D}_{\mathrm{buf}}\) for the replay buffer and \(\rho_{\mathcal{D}}\) for its state marginal.

The actor is a diagonal tanh-squashed Gaussian policy \(\pi_\theta(a\mid s)\), with ensemble size \(N_Q\), index set \(\mathcal I_Q=\{1,\ldots,N_Q\}\), critics \(Q_i\), target critics \(\bar Q_i\), and temperature \(\alpha\). Entropy, conservative-learning, and fixed-policy-backup notation are defined where used.

\section{Sigmoid-Bounded Conservative Q-Learning}
\label{sec:method}

\paragraph{SCQ Overview.}
SCQ combines three components. First, SigEnt replaces the sample-wise entropy contribution in both actor and critic updates with a strictly positive, bounded score and tunes its coefficient to a bounded target entropy. Second, a Cal-QL-style pessimistic critic ensemble combines the entropy-augmented TD objective with conservative regularization and return-based calibration on actions sampled from the current policy. Third, critic-only LayerNorm stabilizes value optimization while leaving the actor architecture unchanged. Algorithmic details and the training-dynamics diagnostic are given below.

\subsection{Sigmoid-Bounded Entropy Mechanism}

We consider a tanh-squashed Gaussian policy $a=\tanh(x)$, where $x$ is sampled via reparameterization from the policy network outputs: $x=\mu_\theta(s)+\sigma_\theta(s)\odot\epsilon$, with $\epsilon\sim\mathcal{N}(0,I)$. We express all environment actions in the common normalized coordinates $a\in[-1,1]^{d_a}$; environment-specific box bounds are affinely mapped to this interface. Let $\log \pi_{\theta,i}(a_i|s)$ denote the \emph{per-dimension} log-density of the squashed policy, so that $\log \pi_\theta(a|s)=\sum_{i=1}^{d_a}\log \pi_{\theta,i}(a_i|s)$. We define the per-dimension surprisal as $\ell_i=-\log \pi_{\theta,i}(a_i|s)$. The bounded entropy contribution is computed with the logistic sigmoid $\operatorname{sigmoid}(u)=(1+e^{-u})^{-1}$:

\begin{equation}
\label{eq:sigmoid_entropy}
\begin{aligned}
h_i(\ell_i)
&=h_{\max}\cdot \operatorname{sigmoid}\!\left(\frac{\ell_i-m}{t}\right), \\
\mathcal{H}_{\mathrm{sig}}(s,a;\theta)
&=\sum_{i=1}^{d_a} h_i(\ell_i).
\end{aligned}
\end{equation}
By construction, $\mathcal{H}_{\mathrm{sig}}(s,a;\theta)\in(0,d_a h_{\max})$.

\subsection{Actor and Temperature Learning}

The actor minimizes the entropy-regularized objective
\begin{equation}
\label{eq:actor_loss_sig}
\mathcal{L}_{\pi}
=
\Expectation{s\sim\rho_{\mathcal{D}},\,a\sim\pi_\theta(\cdot\mid s)}
\left[
-\min_{i\in\mathcal{I}_Q}Q_i(s,a)
-\alpha\mathcal{H}_{\mathrm{sig}}(s,a;\theta)
\right].
\end{equation}
We tune $\alpha>0$ to match the bounded-entropy target
\begin{equation}
\label{eq:temperature_constraint_sig}
\Expectation{s\sim\rho_{\mathcal{D}},\,a\sim\pi_\theta(\cdot\mid s)}
\left[\mathcal{H}_{\mathrm{sig}}(s,a;\theta)\right]
=\mathcal{H}_{\mathrm{target}}.
\end{equation}
With the shared target pre-squash standard deviation $\sigma_{\mathrm{tar}}=0.1$, we set $\mathcal{H}_{\mathrm{target}}=d_a h(\bar\ell(0.1))\approx0.25743d_a$, where $\bar\ell$ is defined in Eq.~\ref{eq:shared_shape_sensitivity}. This target instantiates automatic temperature tuning; it does not select or modify the SigEnt shape.

\subsection{Critic Learning}

\paragraph{Critic Loss Function.}

The entropy-augmented Bellman target is defined as:
\begin{equation}
\label{eq:bellman_target_sig}
y = r + (1-d_{\mathrm{term}})\gamma \left( \min_{i\in\mathcal{I}_Q} \bar Q_i(s', a') + \alpha\, \mathcal{H}_{\mathrm{sig}}(s',a';\theta) \right),
\end{equation}
where $a' \sim \pi_\theta(\cdot|s')$ and $\bar Q_i$ denotes the target critic. The base TD loss is
\begin{equation}
\label{eq:td_loss}
\mathcal{L}_{\mathrm{TD}}^{(i)} = \Expectation{\mathcal{D}_{\mathrm{buf}}} \Big[\big(Q_i(s,a) - y\big)^2 \Big], \qquad i\in\mathcal{I}_Q.
\end{equation}

We add a simplified CQL-style conservative regularizer~\cite{cql}, unchanged from Cal-QL~\cite{calql}: policy-sampled actions from both $s$ and $s'$ are calibrated against the Monte-Carlo return as a lower bound, then penalized through a log-sum-exp gap against the buffer action's value. This yields a conservative loss $\mathcal{L}_{\mathrm{CQL}}^{(i)}$ and the final critic objective
\begin{equation}
\label{eq:critic_final_loss}
\mathcal{L}_{Q_i}
=
\mathcal{L}_{\mathrm{TD}}^{(i)}
+
\lambda_{\mathrm{cql}}\,\mathcal{L}_{\mathrm{CQL}}^{(i)},
\qquad i\in\mathcal{I}_Q,
\end{equation}
where $\lambda_{\mathrm{cql}}\ge0$ controls the strength of the conservative regularization. The full derivation, including the policy-sampled action set, return-based calibration, and log-sum-exp regularizer, is given in Appendix~\ref{app:critic_loss_details}.

\paragraph{Critic-only LayerNorm.}
Following RLPD~\cite{rlpd}, SCQ applies LayerNorm only in the critic networks used in Eqs.~\ref{eq:td_loss}--\ref{eq:critic_final_loss}. The actor architecture is unchanged, so this normalization directly targets critic optimization without introducing an additional policy regularizer.

\begin{table*}[t]
\centering
\caption{Full-demo comparison of SCQ with Cal-QL, RLPD, and FlashSAC. We screened SCQ, Cal-QL, CQL, IQL, AWAC, RLPD, SAC+OD, and FlashSAC, then selected the strongest and  most relevant available comparators. Entries report endpoint means and sample  standard deviations across the evaluated seeds. We report Cal-QL with its standard pretraining because a no-pretraining variant performed substantially weaker in our pilot, and method-faithful pretraining is needed for a competitive baseline. The aggregate row reports the mean $\pm$ sample standard deviation across the 13 displayed task-level means. ``--'' marks an unavailable result.}
\label{tab:full_demo_results}
\begin{tabular*}{\textwidth}{@{\extracolsep{\fill}}llcccc@{}}
\toprule
Family & Dataset variant & SCQ & Cal-QL & RLPD & FlashSAC \\
\midrule
AntMaze & medium-play-v2 & 0.58 $\pm$ 0.19 & 0.43 $\pm$ 0.06 & 0.40 $\pm$ 0.26 & $\mathbf{0.60 \pm 0.16}$ \\
AntMaze & medium-diverse-v2 & 0.26 $\pm$ 0.09 & 0.20 $\pm$ 0.10 & 0.07 $\pm$ 0.06 & $\mathbf{0.38 \pm 0.15}$ \\
AntMaze & large-play-v2 & 0.20 $\pm$ 0.07 & $\mathbf{0.23 \pm 0.12}$ & 0.03 $\pm$ 0.06 & 0.18 $\pm$ 0.11 \\
AntMaze & large-diverse-v2 & 0.18 $\pm$ 0.11 & $\mathbf{0.23 \pm 0.15}$ & 0.00 $\pm$ 0.00 & 0.22 $\pm$ 0.22 \\
Adroit Pen & human-v2 & $\mathbf{1.00 \pm 0.00}$ & 0.77 $\pm$ 0.23 & 0.87 $\pm$ 0.15 & 0.93 $\pm$ 0.06 \\
Adroit Pen & cloned-v2 & $\mathbf{1.00 \pm 0.00}$ & 0.60 $\pm$ 0.10 & 0.90 $\pm$ 0.10 & $\mathbf{1.00 \pm 0.00}$ \\
Adroit Pen & expert-v2 & $\mathbf{1.00 \pm 0.00}$ & 0.80 $\pm$ 0.10 & 0.80 $\pm$ 0.10 & $\mathbf{1.00 \pm 0.00}$ \\
Adroit Relocate & human-v2 & $\mathbf{0.95 \pm 0.07}$ & 0.00 $\pm$ 0.00 & 0.77 $\pm$ 0.15 & 0.00 $\pm$ 0.00 \\
Adroit Relocate & cloned-v2 & $\mathbf{0.95 \pm 0.07}$ & 0.00 $\pm$ 0.00 & 0.53 $\pm$ 0.47 & 0.00 $\pm$ 0.00 \\
Adroit Relocate & expert-v2 & $\mathbf{1.00 \pm 0.00}$ & 0.00 $\pm$ 0.00 & 0.90 $\pm$ 0.10 & 0.87 $\pm$ 0.12 \\
Kitchen & complete-v2 & $\mathbf{1.00 \pm 0.00}$ & 0.00 $\pm$ 0.00 & $\mathbf{1.00 \pm 0.00}$ & 0.67 $\pm$ 0.58 \\
Kitchen & partial-v2 & $\mathbf{1.00 \pm 0.00}$ & 0.00 $\pm$ 0.00 & $\mathbf{1.00 \pm 0.00}$ & $\mathbf{1.00 \pm 0.00}$ \\
Kitchen & mixed-v2 & $\mathbf{1.00 \pm 0.00}$ & 0.00 $\pm$ 0.00 & 0.33 $\pm$ 0.58 & 0.00 $\pm$ 0.00 \\
\midrule
Aggregate & Macro mean (13 tasks) & $\mathbf{0.78 \pm 0.34}$ & 0.25 $\pm$ 0.30 & 0.58 $\pm$ 0.38 & 0.53 $\pm$ 0.41 \\
\bottomrule

\end{tabular*}
\end{table*}

\begin{figure*}[t]
\centering
\includegraphics[width=\textwidth]{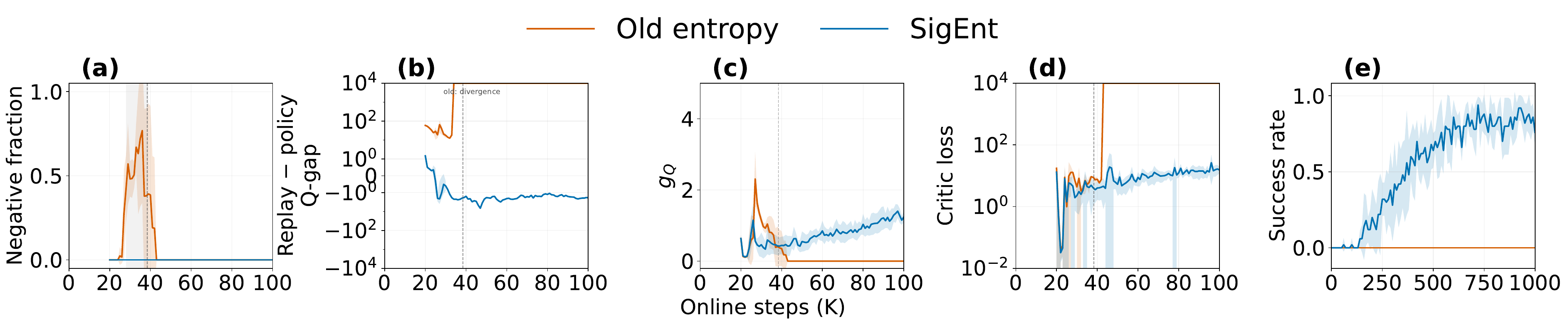}
\caption{Failure anatomy under the frozen RLPD Relocate-Cloned protocol. Each panel reports five-seed means with $\pm$ one sample standard deviation for old log-probability entropy and SigEnt, using the same replay mixture, critic backup, and 1M-step budget. We use the default per-dimension targets: $-1$ for old log-probability entropy and target standard deviation $0.1$ for SigEnt. The panels show negative-entropy fraction, replay--policy effective-$Q$ gap, local $g_Q$ sensitivity, critic loss, and evaluation success. Old entropy develops a substantial negative-entropy fraction around 28--39K steps, followed by a reduced effective-$Q$ gap, critic divergence, and zero success. SigEnt remains nonnegative and finite, with stable critic training and nonzero success. For readability, Q-gap and critic-loss values exceeding the symmetric display bound of $10^4$ are clipped; Q-gap uncertainty shading is omitted after 30K steps once the dispersion becomes dominated by divergent runs. This failure anatomy measures the temporal sequence of policy-update and value-estimation instability; the direct score intervention is provided separately by the clipped log-probability comparison in Table~\ref{tab:positive_entropy_comparison}.}
\label{fig:relocate_failure_anatomy}
\end{figure*}

\subsection{Theoretical Properties}

The critic target in Eq.~\ref{eq:bellman_target_sig} motivates an idealized fixed-policy backup in which expectations are exact and target-network lag is omitted. For compactness, let $m_{\mathbf Q}(s,a)=\min_{i\in\mathcal{I}_Q}Q_i(s,a)$ and define
\begin{equation}
\label{eq:sigent_backup_value}
V_{\mathbf Q}^{\pi_\theta,\alpha}(s)
=\Expectation{a\sim\pi_\theta(\cdot\mid s)}
\left[m_{\mathbf Q}(s,a)+\alpha\mathcal{H}_{\mathrm{sig}}(s,a;\theta)\right],
\end{equation}
\begin{equation}
\label{eq:sigent_scalar_backup}
\mathcal{B}_{\mathbf Q}^{\pi_\theta,\alpha}(s,a)
=\Expectation{r,s',d_{\mathrm{term}}}
\left[r+(1-d_{\mathrm{term}})\gamma V_{\mathbf Q}^{\pi_\theta,\alpha}(s')\right].
\end{equation}
\begin{equation}
\label{eq:sigent_backup_operator}
\left[\mathcal{T}_{\mathrm{sig}}^{\pi_\theta}\mathbf Q\right]_i(s,a)
=\mathcal{B}_{\mathbf Q}^{\pi_\theta,\alpha}(s,a),
\quad i\in\mathcal{I}_Q.
\end{equation}
The following results establish contraction and entropy-scale equivariance for this idealized setting; they do not assert convergence of the complete neural actor--critic training dynamics.

\begin{theorem}[Idealized SigEnt Backup]
\label{thm:sigent_contraction}
Assume bounded rewards, and fix any finite $N_Q\ge2$, $\pi_\theta$, and $\alpha>0$. On the space of bounded critic ensembles with norm $\|\mathbf Q\|_\infty=\max_{i\in\mathcal{I}_Q}\|Q_i\|_\infty$, the operator in Eq.~\ref{eq:sigent_backup_operator} is a $\gamma$-contraction: for any $\mathbf Q$ and $\widetilde{\mathbf Q}$,
\begin{equation}
\label{eq:sigent_contraction}
\left\|
\mathcal{T}_{\mathrm{sig}}^{\pi_\theta}\mathbf{Q}
-\mathcal{T}_{\mathrm{sig}}^{\pi_\theta}\widetilde{\mathbf{Q}}
\right\|_\infty
\le
\gamma\left\|\mathbf{Q}-\widetilde{\mathbf{Q}}\right\|_\infty.
\end{equation}
Consequently, the idealized fixed-policy backup has a unique fixed point and its exact Q-iteration converges to that fixed point.
\end{theorem}

\paragraph{Proof sketch.}
The entropy term is independent of $\mathbf{Q}$ and therefore cancels when comparing two backups. Moreover, the finite-ensemble minimum is non-expansive:
\begin{equation}
\label{eq:critic_ensemble_nonexpansive}
\left|m_{\mathbf Q}(s',a')-m_{\widetilde{\mathbf Q}}(s',a')\right|
\le \left\|\mathbf Q-\widetilde{\mathbf Q}\right\|_\infty.
\end{equation}
Taking the expectation in Eq.~\ref{eq:sigent_backup_operator} gives Eq.~\ref{eq:sigent_contraction}; the fixed-point claim follows from the Banach fixed-point theorem. A complete proof is provided in Appendix~\ref{app:sigent_contraction}.

\begin{theorem}[$\alpha\mathcal{H}$ equilibrium equivariance]
\label{thm:alpha_h_equivariance}
Suppose that the critic objective depends on entropy magnitude $h_{\max}$ and temperature $\alpha$ only through $\alpha\mathcal{H}_{\mathrm{sig}}$ in its Bellman target, while all other critic terms and sampling distributions remain unchanged. Let $(Q^*,\theta^*,\alpha^*)$ with $\alpha^*>0$ be a stationary equilibrium of the critic objective, Eq.~\ref{eq:actor_loss_sig}, and Eq.~\ref{eq:temperature_constraint_sig}. For any $c>0$, define
\begin{equation}
\label{eq:alpha_h_rescaling}
h_{\max}'=\frac{h_{\max}}{c},
\qquad
\mathcal{H}_{\mathrm{target}}'
=\frac{\mathcal{H}_{\mathrm{target}}}{c},
\qquad
\alpha'=c\alpha^*.
\end{equation}
Then $(Q^*,\theta^*,\alpha')$ is a stationary equilibrium of the rescaled system, and
\begin{equation}
\label{eq:alpha_h_invariance}
\alpha'\mathcal{H}_{\mathrm{sig}}'
=\alpha^*\mathcal{H}_{\mathrm{sig}}.
\end{equation}
\end{theorem}

Theorem~\ref{thm:alpha_h_equivariance} shows that $h_{\max}$ is not an independent equilibrium-scale parameter when $\mathcal{H}_{\mathrm{target}}$ is rescaled jointly. It does not imply that finite training trajectories are invariant under this transformation. The complete proof is provided in Appendix~\ref{app:alpha_h_equivariance}.

\paragraph{Shared action-normalized shape parameterization.}
Theorem~\ref{thm:alpha_h_equivariance} motivates the canonical score normalization $h_{\max}=1$. We fix the per-dimension SigEnt shape $(m,t)=(-0.3,0.55)$ once in the common normalized action coordinates. The design criterion is to place the practical policy-spread regime in the responsive region of the sigmoid score, rather than in either saturated tail. Under the near-zero tanh-Jacobian approximation,
\begin{equation}
\label{eq:shared_shape_sensitivity}
\begin{aligned}
\bar\ell(\sigma)
&=\log\sigma+\tfrac12\bigl[\log(2\pi)+1\bigr]
  +\log(1+10^{-3}), \\
p
&=\operatorname{sigmoid}\!\left(\frac{\ell-m}{t}\right), \\
\frac{\partial h}{\partial\ell}
&=\frac{h_{\max}}{t}p(1-p).
\end{aligned}
\end{equation}
For the fixed shape, $\sigma\in[0.105,0.304]$ retains at least $80\%$ of the maximal score derivative with respect to surprisal (and, under this local approximation, log-standard-deviation). The shared $\sigma_{\mathrm{tar}}=0.1$ lies within this high-sensitivity interval and is used only to instantiate the automatic-temperature target. The per-dimension shape depends on the shared normalized action parameterization and fixed responsiveness convention, not task dynamics, rewards, horizons, datasets, returns, or physical action units. It is unchanged when $d_a$ varies; action dimension affects only the summed target entropy.

\begin{figure*}[!t]
  \centering
  \includegraphics[width=\textwidth]{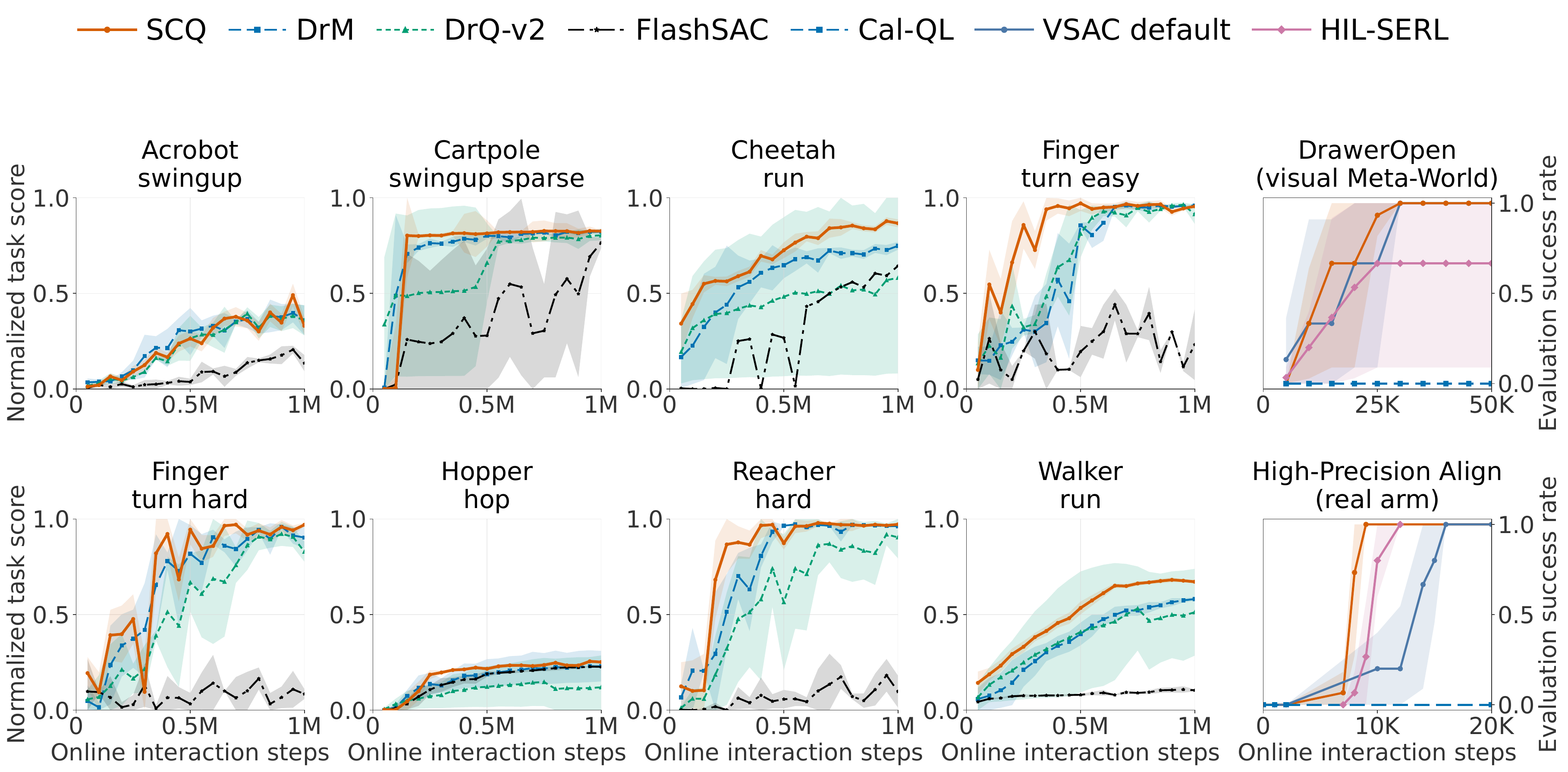}
  \caption{Cross-domain visual-control results. The first four columns show Visual DMC evaluation curves over 1M decision steps. Baseline bands show $\pm$ one sample standard deviation. The right column shows success-rate means with $\pm$ sample-standard-deviation bands for DrawerOpen (visual Meta-World, 50K horizon) and High-Precision Align (real arm, 20K horizon). HIL-SERL uses the matched four-frame DrM visual backbone without human intervention; its Align curve ends at the last recorded 12K evaluation. Cal-QL remains at zero success rate in both tasks. VSAC is the standard tanh-Gaussian SAC reference with automatic target entropy $-d_a$ and log-probability entropy in both actor and critic backup; SCQ uses target standard deviation 0.1.}
  \label{fig:visual_sota_results}
\end{figure*}

\begin{figure*}[!t]
  \centering
  \includegraphics[width=\textwidth]{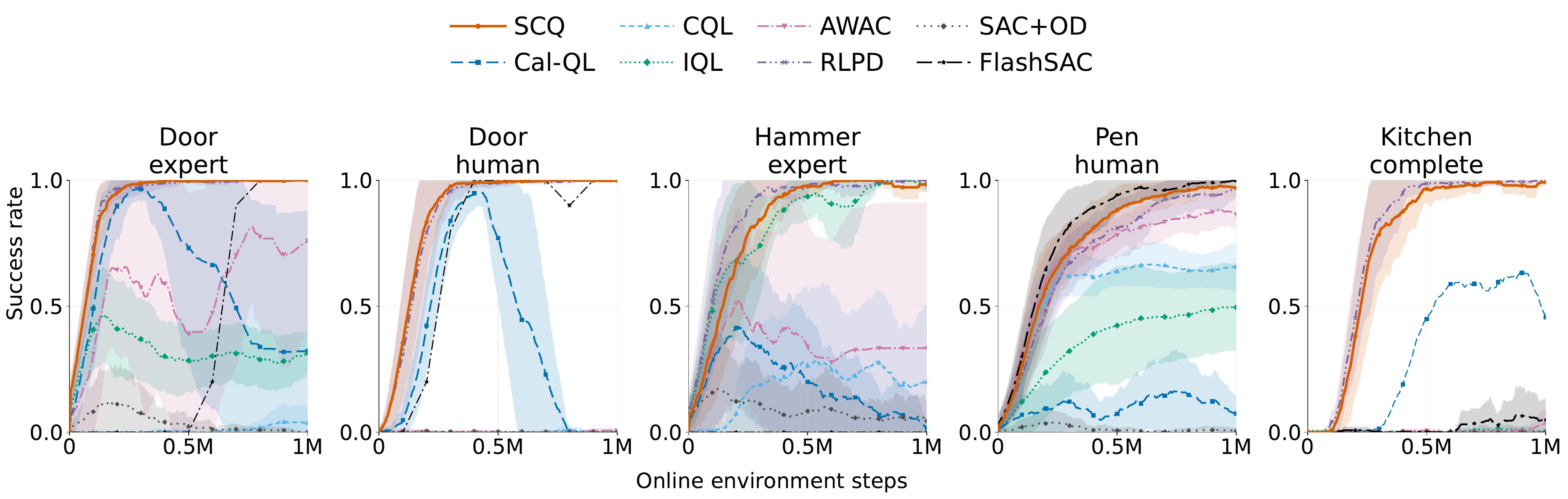}
  \caption{One-shot comparison over 1M online steps from one successful demonstration. Curves show the mean over evaluated seeds, with shaded regions indicating $\pm$ one sample standard deviation. All curves are display-anchored at zero online steps; this anchor is not an evaluation.}
  \label{fig:sim_benchmarks}
\end{figure*}

\section{Experiments}
\label{sec:experiments}

\subsection{Experimental Protocol}

We study five questions: \textbf{Q1}, whether replacing the standard log-probability entropy with a calibrated positive entropy score improves evaluation success and online sample efficiency; \textbf{Q2}, whether the complete SCQ design generalizes to standard full-dataset offline-to-online benchmarks; \textbf{Q3}, whether SCQ is competitive with visual RL baselines under matched image-based inputs; \textbf{Q4}, whether SCQ remains data-efficient when only one successful demonstration is available; and \textbf{Q5}, whether SCQ can be deployed across real-robot embodiments and, where a common reference exists, reduce within-task completion steps.

\paragraph{Common protocol.}
Formal simulation comparisons use common offline data, online budget, evaluation schedule, and checkpoint rule; results report seed means and standard deviations.

\paragraph{Baselines.}
State-based comparisons include Cal-QL, CQL~\cite{calql,cql}, RLPD~\cite{rlpd}, AWAC~\cite{awac}, IQL~\cite{iql}, SAC+OD, and FlashSAC~\cite{kim2026flashsac}. FlashSAC is the most recent strong offline-to-online baseline available to us and the RSS 2026 Best Paper. We screened the mainstream Minari/D4RL task families and selected Cal-QL, RLPD, and FlashSAC as the strongest and most relevant comparators available in our evaluation environment, not as a performance-dependent subset. Visual comparisons use SCQ, DrM~\cite{xu2024drm}, DrQ-v2~\cite{yarats2022drqv2}, and FlashSAC with matched inputs and interaction budgets. 

\paragraph{Benchmark suites and metrics.}
The full-dataset suite has 13 tasks (four AntMaze, three Pen, three Relocate, and three Kitchen); the visual suite has eight pre-specified DMC tasks. We report normalized return or success rate, online sample efficiency, and final performance under frozen task contracts.

\subsubsection{Shared Action-Normalized SigEnt Configuration}

All visual, full-demo, and one-shot policies use the common tanh-squashed $[-1,1]$ action interface. Table~\ref{tab:reference_variance_audit} records the single shared SigEnt configuration; no task-specific action-range, shape, or target override is applied.

\begin{table}[!t]
\centering
\normalsize
\caption{Shared action-normalized SigEnt configuration for all formal SCQ protocols.}
\label{tab:reference_variance_audit}
\begin{tabular*}{\columnwidth}{@{\extracolsep{\fill}}lcccc@{}}
\toprule
Protocol & $\sigma_{\mathrm{tar}}$ & $h_{\max}$ & $m$ & $t$ \\
\midrule
All formal protocols & 0.1 & 1 & $-0.3$ & 0.55 \\
\bottomrule
\end{tabular*}
\end{table}

\begin{figure*}[!t]
  \centering
  \includegraphics[width=\textwidth]{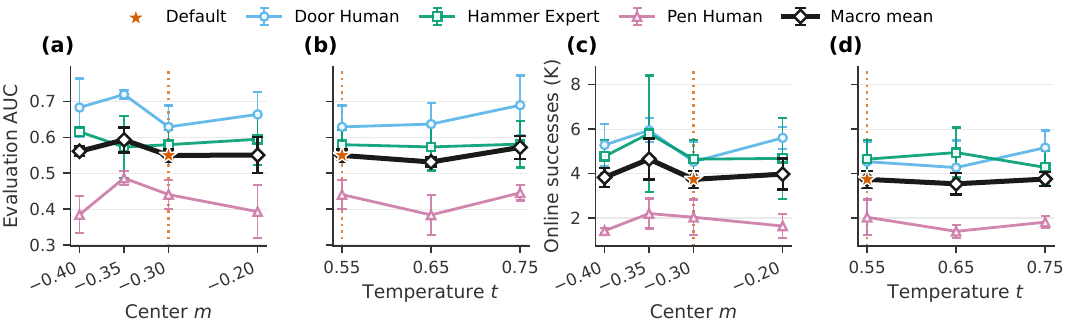}
  \caption{Sensitivity to the shared SigEnt center $m$ and temperature $t$ under the frozen 400K one-shot protocol. Curves show task means for Door Human, Hammer Expert, and Pen Human; error bars are sample standard deviations across three seeds. The thick black curve is the macro mean computed per seed across the three tasks, and the orange star marks the pre-specified default $(m,t)=(-0.3,0.55)$. The tested neighborhood shows no catastrophic degradation or consistent monotonic trend; the default remains within the robust region rather than being chosen by task-specific tuning.}
  \label{fig:sigent_mt_sensitivity}
\end{figure*}

We additionally evaluate the fixed SigEnt shape in a local center/temperature sweep while keeping the target standard deviation at $0.1$. The sweep uses six configurations around the pre-specified default $(m,t)=(-0.3,0.55)$ on Door Human, Hammer Expert, and Pen Human, with three seeds per task. Figure~\ref{fig:sigent_mt_sensitivity} reports evaluation AUC and online-success counts. Performance remains within a stable neighborhood without a consistent monotonic trend; different tasks prefer different nearby settings, and the default is not selected post hoc from this sweep.

\subsection{Mechanism Validation}

\subsubsection{Positive-Entropy Test (Q1)}

We test whether a calibrated positive entropy score improves one-shot learning relative to the standard log-probability entropy. All conditions use the same actor, critic, replay, calibration, and evaluation protocol. The direct intervention $H_{\mathrm{clip}}=\sum_i\max(0,-\log\pi_i)$ preserves the standard score wherever it is positive and removes only its negative branch; its temperature target is calibrated to the same reference policy standard deviation, $\sigma_{\mathrm{tar}}=0.1$. SigEnt, matched ReLU, and matched Softplus are matched in target value and local actor-gradient scale at this reference point.

\begin{table}[!t]
\centering
\caption{Entropy-form comparison under the frozen one-shot protocol. Clipped log-probability is the direct intervention $H_{\mathrm{clip}}=\sum_i\max(0,-\log\pi_i)$: it preserves the positive branch of the standard score and removes only negative contributions, with its temperature target calibrated to the same reference policy standard deviation. Matched ReLU and Softplus are locally first-order matched to SigEnt. Entries report task-level means $\pm$ sample standard deviations; task-mean uncertainties use independent-error propagation across the task-level standard deviations. ``First 100\%'' is the earliest 10/10 evaluation.}
\label{tab:positive_entropy_comparison}
\setlength{\tabcolsep}{2pt}
\renewcommand{\arraystretch}{1.08}
\scalebox{0.9}{%
\begin{tabular}{llccc}
\toprule
Task & Score & First 100\% & AUC $\uparrow$ & Online succ. $\uparrow$ \\
\midrule
Door & Std. log-prob. & $142.8 \pm 17.5$K & $0.647 \pm 0.001$ & $4{,}323 \pm 88$ \\
Door & SigEnt & $147.8 \pm 28.2$K & $0.663 \pm 0.151$ & $5{,}551 \pm 647$ \\
Door & Matched ReLU & $\mathbf{133.6 \pm 35.5}$K & $\mathbf{0.694 \pm 0.069}$ & $\mathbf{5{,}713 \pm 946}$ \\
Door & Softplus & $156.8 \pm 7.6$K & $0.670 \pm 0.034$ & $5{,}220 \pm 446$ \\
Door & Clipped log-prob. & $153.5 \pm 36.7$K & $0.656 \pm 0.012$ & $5{,}451 \pm 508$ \\
\midrule
Hammer & Std. log-prob. & $165.1 \pm 28.4$K & $0.437 \pm 0.057$ & $3{,}358 \pm 709$ \\
Hammer & SigEnt & $\mathbf{105.0 \pm 49.5}$K & $\mathbf{0.571 \pm 0.078}$ & $\mathbf{5{,}260 \pm 2{,}086}$ \\
Hammer & Matched ReLU & $160.1 \pm 47.6$K & $0.552 \pm 0.050$ & $5{,}051 \pm 1{,}181$ \\
Hammer & Softplus & $168.5 \pm 58.0$K & $0.555 \pm 0.053$ & $4{,}566 \pm 824$ \\
Hammer & Clipped log-prob. & $173.4 \pm 31.7$K & $0.513 \pm 0.085$ & $4{,}389 \pm 727$ \\
\midrule
Pen & Std. log-prob. & $308.1 \pm 33.0$K & $0.372 \pm 0.001$ & $1{,}536 \pm 74$ \\
Pen & SigEnt & $\mathbf{288.2 \pm 27.0}$K & $\mathbf{0.509 \pm 0.060}$ & $\mathbf{2{,}086 \pm 87}$ \\
Pen & Matched ReLU & $322.0 \pm 49.0$K & $0.422 \pm 0.100$ & $1{,}641 \pm 389$ \\
Pen & Softplus & $317.6 \pm 17.6$K & $0.483 \pm 0.069$ & $2{,}051 \pm 619$ \\
Pen & Clipped log-prob. & $--$ & $0.425 \pm 0.117$ & $1{,}858 \pm 574$ \\
\midrule
\multirow{5}{*}{Task mean} & Std. log-prob. & $205.3\!\pm\!15.6$K & $0.485\!\pm\!0.019$ & $3{,}072\!\pm\!239$ \\
 & SigEnt & $\mathbf{180.3\!\pm\!29.2}$K & $\mathbf{0.581\!\pm\!0.060}$ & $\mathbf{4{,}299\!\pm\!1{,}124}$ \\
 & Matched ReLU & $205.2\!\pm\!25.7$K & $0.556\!\pm\!0.044$ & $4{,}135\!\pm\!521$ \\
 & Softplus & $214.3\!\pm\!20.4$K & $0.569\!\pm\!0.031$ & $3{,}946\!\pm\!374$ \\
 & Clipped log-prob. & $--$ & $0.532\!\pm\!0.049$ & $3{,}899\!\pm\!352$ \\
\bottomrule
\end{tabular}
}
\end{table}

Table~\ref{tab:positive_entropy_comparison} first provides a direct intervention on the sign of the entropy contribution. Clipped log-probability improves over standard entropy on every task, increasing task-mean AUC from 0.485 to 0.532 and online successes from 3,072 to 3,899. Because its score transformation preserves the standard score wherever it is positive and changes only the negative branch, the result directly supports the empirical role of entropy positivity. SigEnt is stronger still, reaching a task-mean AUC of 0.581 and 4,299 online successes. Matched ReLU and Softplus show the same directional improvement over standard entropy on both task-mean quantities, with AUCs of 0.556 and 0.569 and online-success counts of 4,135 and 3,946. Together, these controls identify removal of negative contributions as an important driver, while the smooth bounded SigEnt shape offers an additional aggregate benefit. Pen's clipped First 100\% entry is ``--'' because two of its five seeds never reached a 10/10 evaluation.

The clipped comparison is the primary intervention evidence for the positivity hypothesis. Negative sample-wise entropy changes the local effective-Q landscape and the policy-mean update; the moving policy then changes the critic's queried actions, and the resulting value-estimate changes feed back into subsequent policy updates. Figure~\ref{fig:entropy_mechanism_gradient} provides complementary training-dynamics evidence for this feedback through the temporal variability of $g_Q$.

\subsubsection{Component-Interaction Ablation}

We ablate the inherited Cal-QL calibration and critic LayerNorm under the same 400K-step protocol. Each entry aggregates five independent seeds.

\begin{table}[!t]
\centering
\caption{Component-interaction ablation on Door Human, Hammer Expert, and Pen Human at 400K one-shot steps. Task-level entries are mean $\pm$ sample standard deviation over five seeds; task-mean uncertainties use independent-error propagation. AUC is normalized by the 400K-step horizon; Succ. counts successful online training episodes in thousands. LN denotes critic LayerNorm and Cal. denotes Cal-QL calibration.}
\label{tab:component_ablation}
\footnotesize
\setlength{\tabcolsep}{1.5pt}
\renewcommand{\arraystretch}{1.08}
\begin{tabular*}{\columnwidth}{@{\extracolsep{\fill}}llcccc@{}}
\toprule
Task & Metric & SCQ & \shortstack{Std.\\log-prob.} & w/o LN & w/o Cal. \\
\midrule
\multirow{2}{*}{Door} & AUC $\uparrow$ & $0.67\!\pm\!0.11$ & $0.66\!\pm\!0.03$ & $0.60\!\pm\!0.11$ & $\mathbf{0.68\!\pm\!0.07}$ \\
 & Succ. (K) $\uparrow$ & $\mathbf{5.43\!\pm\!1.88}$ & $4.90\!\pm\!1.00$ & $4.64\!\pm\!0.56$ & $5.13\!\pm\!0.90$ \\
\midrule
\multirow{2}{*}{Hammer} & AUC $\uparrow$ & $\mathbf{0.60\!\pm\!0.07}$ & $0.44\!\pm\!0.04$ & $0.52\!\pm\!0.19$ & $0.55\!\pm\!0.10$ \\
 & Succ. (K) $\uparrow$ & $\mathbf{5.86\!\pm\!1.80}$ & $3.46\!\pm\!0.53$ & $3.82\!\pm\!2.98$ & $3.81\!\pm\!1.40$ \\
\midrule
\multirow{2}{*}{Pen} & AUC $\uparrow$ & $\mathbf{0.47\!\pm\!0.08}$ & $0.42\!\pm\!0.09$ & $0.25\!\pm\!0.16$ & $0.46\!\pm\!0.11$ \\
 & Succ. (K) $\uparrow$ & $1.88\!\pm\!0.36$ & $1.96\!\pm\!0.73$ & $1.05\!\pm\!0.72$ & $\mathbf{2.14\!\pm\!1.14}$ \\
\midrule
Mean & AUC $\uparrow$ & $\mathbf{0.58\!\pm\!0.05}$ & $0.51\!\pm\!0.03$ & $0.46\!\pm\!0.09$ & $0.56\!\pm\!0.05$ \\
 & Succ. (K) $\uparrow$ & $\mathbf{4.39\!\pm\!0.88}$ & $3.44\!\pm\!0.45$ & $3.17\!\pm\!1.04$ & $3.69\!\pm\!0.67$ \\
\bottomrule
\end{tabular*}
\end{table}

Table~\ref{tab:component_ablation} shows that full SCQ has the highest task-mean AUC (0.579) and online-success count (4.39K). Replacing SigEnt with standard log-probability entropy reduces the mean AUC, driven by Hammer and Pen; removing critic LayerNorm reduces both aggregate measures and is particularly harmful on Pen. Calibration is task-dependent: removing it is marginally stronger on Door Human, but full SCQ is higher on Hammer and Pen and leads the task mean. Thus, calibration is retained as a robust default rather than claimed as a uniformly dominant component.

\subsection{Generalization Benchmarks}

\subsubsection{Standard Full-Dataset Generalization (Q2)}

Table~\ref{tab:full_demo_results} compares SCQ with Cal-QL, RLPD, and FlashSAC across 13 full-demo tasks under audited common-budget protocols. Entries report means and sample standard deviations.
SCQ does not uniformly dominate on AntMaze: FlashSAC is stronger on the medium variants, and Cal-QL is comparable on the large variants. The aggregate margin is instead driven largely by Adroit and Kitchen, where Cal-QL collapses to near-zero performance. We attribute this collapse to a baseline-implementation factor rather than to the entropy formulation under study: the released Cal-QL critic does not apply LayerNorm, whereas SCQ and RLPD both use critic-only LayerNorm (Sec.~\ref{sec:method}), which is known to stabilize offline-to-online critic training and whose absence is a plausible driver of the mid-training collapse observed here. This LayerNorm-driven explanation is orthogonal to the entropy-formulation contribution: the component-interaction ablation (Table~\ref{tab:component_ablation}) isolates SigEnt and critic LayerNorm independently and shows both contribute additively rather than either factor alone accounting for the full margin — removing SigEnt alone (keeping LayerNorm) reduces task-mean AUC from $0.58$ to $0.51$, and removing LayerNorm alone (keeping SigEnt) reduces it further to $0.46$, so SigEnt's benefit is not merely a proxy for LayerNorm.
To diagnose training-dynamics stability in this case study, we track the mean per-state norm of the direct Q-induced gradient on the policy mean,
\begin{equation}
\label{eq:q_gradient_mu_diagnostic}
g_Q = \mathbb{E}_{s\sim\mathcal{B}}\left[\left\|\nabla_{\mu}\left(-\min_{i\in\mathcal{I}_Q} Q_i\left(s,\tanh(\mu+\sigma\epsilon)\right)\right)\right\|_2\right].
\end{equation}
This metric measures the critic term's immediate update pressure on the policy mean. Intuitively, its temporal second difference behaves like a normalized acceleration of the Q-gradient signal: large swings indicate the critic is pulling the policy mean in an unstable, oscillating direction rather than a smooth one. We bin its logged batch means in 10K-step intervals and report this signed normalized temporal second difference.

Figure~\ref{fig:relocate_failure_anatomy} provides a complementary failure-anatomy study under the frozen RLPD Relocate-Cloned protocol. Both conditions use the same replay mixture, critic-backup entropy, network architecture, optimization schedule, and five seeds; they use their respective default entropy formulations and controller targets. The old log-probability condition uses the default target of $-1$ per action dimension, while SigEnt uses the default target standard deviation $0.1$ per dimension. Accordingly, this case study exposes the temporal failure sequence, while the clipped comparison in Table~\ref{tab:positive_entropy_comparison} supplies the direct sign intervention.

The old-entropy runs develop a substantial negative-contribution fraction around 28--39K steps, reaching approximately $0.48\pm0.11$ at 30K. Over the same interval, the replay--policy effective-$Q$ gap contracts, local $g_Q$ becomes highly variable, and the critic loss reaches numerical divergence at approximately $38.4\pm3.6$K steps. The old-entropy success rate then remains at zero. In contrast, SigEnt keeps the entropy contribution nonnegative and finite, maintains a bounded local update signal, and reaches approximately $0.76\pm0.09$ success by 1M steps.

\subsubsection{Visual Simulation Generalization (Q3)}

Figure~\ref{fig:visual_sota_results} summarizes visual-control performance across eight Visual DMC tasks and two deployment evaluations. SCQ transfers across locomotion, manipulation, and navigation-style control and reaches higher final performance on most displayed tasks. The right-column VSAC comparison differs jointly in entropy score, critic-backup entropy, and temperature target, so it compares deployed methods rather than isolating the entropy score.

\subsection{Data-Efficient Learning}

\subsubsection{One-Shot Data Efficiency (Q4)}

Each Q4 task starts from one successful demonstration. Figure~\ref{fig:sim_benchmarks} shows multi-seed SCQ evaluations together with baselines

\subsection{Real-Robot Deployment (Q5)}

Our real-world suite spans High-Precision Align, Wheeled Ball-to-Goal, Quadruped Slalom, and Humanoid Navigation-Kick under single-view local observations. Figure~\ref{fig:real_world_generalization} summarizes their settings and outcomes.

The deployment summary is complemented by direct SCQ--VSAC comparisons on all four physical tasks and visual Meta-World DrawerOpen. Figure~\ref{fig:visual_sota_results} reports the learning curves for DrawerOpen and High-Precision Align, while Table~\ref{tab:real_robot_coverage} reports the final trial outcomes across the full deployment suite.

\begin{table*}[t]
\centering
\normalsize
\caption{Visual-manipulation and real-robot evaluation coverage. Entries report successful evaluation trials over total trials, and online budget is the maximum permitted number of environment interactions. Every task includes a direct SCQ--VSAC comparison. HIL-SERL is evaluated on all tasks; its DrawerOpen and Align entries pool the endpoint trials from three seeds. DrQ-v2, DrM, and FlashSAC obtain zero successes on Visual DrawerOpen and High-Precision Align; ``--'' denotes that no corresponding run was performed and is not treated as zero.}
\label{tab:real_robot_coverage}
\setlength{\tabcolsep}{2pt}
\renewcommand{\arraystretch}{1.08}
\begin{tabular*}{\textwidth}{@{\extracolsep{\fill}}lrrrrrrrr@{}}
\toprule
Task & Demos & \shortstack{Online\\budget} & SCQ & VSAC & HIL-SERL & DrQ-v2 & DrM & FlashSAC \\
\midrule
Visual DrawerOpen & 1 & 50K & 10/10 & 10/10 & 20/30 & 0/10 & 0/10 & 0/10 \\
High-Precision Align & 0 & 20K & 10/10 & 10/10 & 15/15 & 0/10 & 0/10 & 0/10 \\
Wheeled Ball-to-Goal & 1 & 30K & 6/10 & 2/10 & 1/10 & -- & -- & -- \\
Quadruped Slalom & 1 & 25K & 8/10 & 4/10 & 4/10 & -- & -- & -- \\
Humanoid Navigation-Kick & 0 & 150K & 10/10 & 3/10 & 4/10 & -- & -- & -- \\
\bottomrule
\end{tabular*}
\end{table*}

At the final evaluation, SCQ reaches 10/10 on Visual DrawerOpen, High-Precision Align, and Humanoid Navigation-Kick, and reaches 6/10 on Ball-to-Goal and 8/10 on Quadruped Slalom. VSAC reaches 10/10 on Visual DrawerOpen and High-Precision Align, but obtains 2/10, 4/10, and 3/10 on the other three physical tasks. HIL-SERL obtains 20/30 pooled trials on DrawerOpen and 15/15 on Align across three seeds, and obtains 1/10 and 4/10 on Ball-to-Goal and Quadruped Slalom, respectively. The physical suite additionally tests sparse-data transfer and cross-embodiment generalization.

The four physical tasks also probe distinct forms of task generalization rather than merely repeating a single deployment setting. High-Precision Align reaches approximately $0.2$\,mm pose accuracy and maintains performance across held-out positions and orientations. Wheeled Ball-to-Goal transfers the learned pushing behavior across substantially different ball and goal locations. Quadruped Slalom requires navigation around randomly placed obstacles and exhibits trajectory adaptation through turning and rerouting. Humanoid Navigation-Kick starts from varying positions, avoids the ball when necessary, and reaches the kick location before executing the contact skill. These evaluations therefore test precision, spatial variation, obstacle-conditioned rerouting, and multi-stage navigation, respectively.

Visual DrawerOpen uses a 50K online-interaction budget. The first three physical tasks are capped at 20K--30K interactions. Humanoid Navigation-Kick uses a 150K budget because the robot must navigate from arbitrary starting positions around obstacles to the designated soccer-ball location before executing the kick.

The completion-step panel compares a human reference and learned SCQ policy under the same inference framework; High-Precision Align has no common reference-step measure.

\subsection{Computational Efficiency}

\begin{table}[t]
\centering
\normalsize
\caption{Computational efficiency on the shared 45D/24D interface at batch size 256. Values are medians on an RTX 4090D; counts exclude target critics.}
\label{tab:efficiency}
\begin{tabular*}{\columnwidth}{@{\extracolsep{\fill}}lcccc@{}}
\toprule
Method & \#Q & Params & Update & Actor \\
 & & (M) & (ms) & (ms) \\
\midrule
AWAC & \textbf{2} & \textbf{0.90} & \textbf{0.62} & \textbf{0.16} \\
IQL & \textbf{2} & 1.18 & 0.97 & 0.20 \\
Cal-QL & \textbf{2} & 1.43 & 4.98 & 0.54 \\
CQL & \textbf{2} & 1.44 & 6.37 & 0.53 \\
RLPD & 10 & 1.67 & 1.31 & 0.20 \\
SAC+OD & \textbf{2} & 0.91 & 1.76 & 0.61 \\
FlashSAC & \textbf{2} & 2.47 & 13.69 & 0.52 \\
\textbf{SCQ (ours)} & \textbf{2} & 1.44 & 7.61 & 0.61 \\
\bottomrule

\end{tabular*}
\end{table}

SCQ retains sub-millisecond actor latency, while two critics increase update cost.

\section{Limitations and Conclusion}
\paragraph{Limitations}
We acknowledge certain limitations in our current study. First, although High-Precision Align uses 4-DoF position control and Humanoid Navigation-Kick uses high-level velocity commands, we do not yet demonstrate full-body, joint-level control. Second, the real-world suite remains limited in scale and horizon. It includes high-precision alignment, but not contact-rich, multi-stage manipulation or long-horizon task composition; moreover, the humanoid result evaluates the navigation skill rather than jointly training the complete velocity--navigation--kick pipeline. Future work will evaluate SCQ under longer-horizon, multi-stage robot learning conditions.

\paragraph{Conclusion}
SCQ replaces standard log-probability entropy with a positive sigmoid-bounded score to limit entropy-induced policy mismatch. The evaluated matched positive scores improve aggregate AUC and online successes over standard entropy. SigEnt remains the default because its bounded score provides the idealized backup guarantee and the highest reported online-success count.


\appendix
\section{Critic Conservative Regularization Details}
\label{app:critic_loss_details}

For each state $s$ in the replay batch, we form a policy-sampled action set by sampling multiple actions from the current policy at both $s$ and $s'$ and concatenating them:
\begin{equation}
\label{eq:policy_sampled_action_set}
\mathcal{A}_{\pi}(s)=\{a^{(j)} \sim \pi_\theta(\cdot|s)\}_{j=1}^{n}\ \cup\ \{a'^{(j)} \sim \pi_\theta(\cdot|s')\}_{j=1}^{n}.
\end{equation}
For these policy-sampled actions, we apply the return-based calibration
\begin{equation}
\label{eq:cql_calibration}
\widetilde{Q}_i(s,\tilde a)=\max\!\left\{Q_i(s,\tilde a),\,G_{\mathrm{MC}}(s)\right\},
\qquad \tilde a\in\mathcal{A}_{\pi}(s).
\end{equation}
Let $\widetilde{\mathcal Q}_i(s)=\{\widetilde Q_i(s,\tilde a):\tilde a\in\mathcal A_{\pi}(s)\}$ collect the calibrated policy-sampled values. For each critic $Q_i$, the conservative loss is
\begin{equation}
\label{eq:cql_cap}
\mathcal{L}_{\mathrm{CQL}}^{(i)}
=\Expectation{(s,a)\sim\mathcal{D}_{\mathrm{buf}}}
\left[
\begin{aligned}
&\tau_{\mathrm{cql}}\log\!\left(\exp\!\left(\tfrac{Q_i(s,a)}{\tau_{\mathrm{cql}}}\right)\right.\\
&\left.\quad+\sum_{\tilde q\in\widetilde{\mathcal Q}_i(s)}\exp\!\left(\tfrac{\tilde q}{\tau_{\mathrm{cql}}}\right)\right)-Q_i(s,a)
\end{aligned}
\right].
\end{equation}
where $\tau_{\mathrm{cql}}>0$ is the log-sum-exp temperature (set to $1$ in our implementation). This term discourages spurious high Q-values on policy-sampled actions and uses the Monte-Carlo return as a lower-bound calibration, following the principle of Cal-QL~\cite{calql}. The final critic objective for each critic $Q_i$ combines this term with the base TD loss as $\mathcal{L}_{Q_i}=\mathcal{L}_{\mathrm{TD}}^{(i)}+\lambda_{\mathrm{cql}}\mathcal{L}_{\mathrm{CQL}}^{(i)}$ (Eq.~\ref{eq:critic_final_loss} in the main text).

\section{Proof of Theorem~\ref{thm:sigent_contraction}}
\label{app:sigent_contraction}

\paragraph{Proof.}
Let $\mathcal{X}$ be the space of bounded finite critic ensembles $\mathbf{Q}=(Q_i)_{i\in\mathcal{I}_Q}$ equipped with the norm $\|\mathbf{Q}\|_\infty=\max_{i\in\mathcal{I}_Q}\|Q_i\|_\infty$. This is a complete normed space. Since $r$ and $\mathcal{H}_{\mathrm{sig}}$ are bounded, Eq.~\ref{eq:sigent_backup_operator} maps $\mathcal{X}$ to itself.

For arbitrary finite collections $\{u_i\}_{i\in\mathcal{I}_Q}$ and $\{v_i\}_{i\in\mathcal{I}_Q}$,
\begin{equation}
\label{eq:appendix_min_nonexpansive}
\left|\min_{i\in\mathcal{I}_Q}u_i-\min_{i\in\mathcal{I}_Q}v_i\right|
\le
\max_{i\in\mathcal{I}_Q}\left|u_i-v_i\right|.
\end{equation}
Applying Eq.~\ref{eq:appendix_min_nonexpansive} pointwise to the two input critic ensembles and using the fact that the entropy contribution is identical for both backups yields
\begin{equation}
\begin{aligned}
&\left|
\mathcal{B}_{\mathbf Q}^{\pi_\theta,\alpha}(s,a)
-\mathcal{B}_{\widetilde{\mathbf Q}}^{\pi_\theta,\alpha}(s,a)
\right| \\
&\quad\le
\gamma\Expectation{s',d_{\mathrm{term}},\,a'}\!\left[
(1-d_{\mathrm{term}})
\left|m_{\mathbf Q}(s',a')-m_{\widetilde{\mathbf Q}}(s',a')\right|
\right] \\
&\quad\le
\gamma\left\|\mathbf{Q}-\widetilde{\mathbf{Q}}\right\|_\infty.
\end{aligned}
\end{equation}
Taking the maximum over $i\in\mathcal{I}_Q$ and the supremum over $(s,a)$ proves Eq.~\ref{eq:sigent_contraction}. Since $\gamma\in(0,1)$ and $\mathcal{X}$ is complete, the Banach fixed-point theorem gives a unique fixed point and convergence of exact Q-iteration. $\square$

\section{Proof of Theorem~\ref{thm:alpha_h_equivariance}}
\label{app:alpha_h_equivariance}

\paragraph{Proof.}
By Eq.~\ref{eq:sigmoid_entropy}, rescaling $h_{\max}$ gives
\begin{equation}
\label{eq:appendix_entropy_rescaling}
\mathcal{H}_{\mathrm{sig}}'(s,a;\theta)
=
\frac{1}{c}\mathcal{H}_{\mathrm{sig}}(s,a;\theta).
\end{equation}
Together with Eq.~\ref{eq:alpha_h_rescaling}, this implies
\begin{equation}
\label{eq:appendix_product_invariance}
\alpha'\mathcal{H}_{\mathrm{sig}}'
=
c\alpha^*\frac{\mathcal{H}_{\mathrm{sig}}}{c}
=
\alpha^*\mathcal{H}_{\mathrm{sig}}.
\end{equation}

The assumed dependence of the critic objective and Eq.~\ref{eq:appendix_product_invariance} give
\begin{equation}
\label{eq:appendix_critic_invariance}
\mathcal{L}_{\mathrm{critic}}^{(h_{\max}/c)}
(Q,\theta,c\alpha)
=
\mathcal{L}_{\mathrm{critic}}^{(h_{\max})}(Q,\theta,\alpha).
\end{equation}
Therefore $\nabla_Q\mathcal{L}_{\mathrm{critic}}^{(h_{\max}/c)}(Q^*,\theta^*,c\alpha^*)=0$.

Similarly, the actor objective satisfies
\begin{equation}
\label{eq:appendix_actor_invariance}
\mathcal{L}_{\pi}^{(h_{\max}/c)}
(\theta,Q^*,c\alpha^*)
=
\mathcal{L}_{\pi}^{(h_{\max})}
(\theta,Q^*,\alpha^*).
\end{equation}
Hence $\nabla_\theta\mathcal{L}_{\pi}^{(h_{\max}/c)}(\theta^*,Q^*,c\alpha^*)=0$. Finally, the rescaled temperature residual is
\begin{equation}
\label{eq:appendix_temperature_invariance}
\mathbb{E}\left[\mathcal{H}_{\mathrm{sig}}'(\cdot;\theta^*)\right]
-\mathcal{H}_{\mathrm{target}}'
=
\frac{1}{c}
\left(
\mathbb{E}\left[\mathcal{H}_{\mathrm{sig}}(\cdot;\theta^*)\right]
-\mathcal{H}_{\mathrm{target}}
\right)
=0.
\end{equation}
All three stationary conditions hold, proving that $(Q^*,\theta^*,c\alpha^*)$ is a stationary equilibrium of the rescaled system. $\square$

\section{Sensitivity Calculation for the Shared SigEnt Shape}
\label{app:shared_shape_sensitivity}

For a pre-squash Gaussian sample $x=\mu+\sigma\epsilon$ with $\epsilon\sim\mathcal{N}(0,1)$, the per-dimension base-distribution surprisal is
\begin{equation}
\ell_{\mathrm{base}}
= \log\sigma+\frac{1}{2}\log(2\pi)+\frac{1}{2}\epsilon^2.
\end{equation}
Consequently, its expectation is $\bar\ell(\sigma)=\log\sigma+\frac{1}{2}\log(2\pi)+\frac{1}{2}+\log(1+\varepsilon_{\mathrm{jac}})$, which is Eq.~\ref{eq:shared_shape_sensitivity} with the near-zero tanh-Jacobian correction. For $h(\ell)=\operatorname{sigmoid}((\ell-m)/t)$ and $p=h(\ell)$, the relative score sensitivity is
\begin{equation}
\frac{\partial h/\partial\ell}{\max_{\ell'}\partial h(\ell')/\partial\ell'}
=4p(1-p).
\end{equation}
Setting this ratio to $\rho$ yields $|z|\leq 2\operatorname{arcosh}(1/\sqrt{\rho})$ for $z=(\ell-m)/t$. With the globally fixed $(m,t)=(-0.3,0.55)$ and $\rho=0.8$, mapping the resulting surprisal interval through $\bar\ell^{-1}$ gives $\sigma\in[0.105,0.304]$. This calculation establishes the high-sensitivity interval reported in the main text; it is a property of the shared normalized per-dimension action parameterization and is not fitted per task.

\bibliographystyle{IEEEtran}
\bibliography{references}

\end{document}